\documentclass{article}
\usepackage{amsmath,graphicx,mlspconf}

\usepackage{subcaption}
\usepackage{amsmath}
\usepackage{amsfonts}
\usepackage{nicefrac}
\usepackage[colorlinks=true]{hyperref}
\usepackage[capitalize,nameinlink]{cleveref}
\usepackage{booktabs,tabularx,array}

\crefname{section}{Sec.}{Secs.}
\crefname{appendix}{App.}{Apps.}

\usepackage[numbers,square,sort&compress]{natbib}

\usepackage{tikz,pgfplots}
\usetikzlibrary{plotmarks,shapes,shapes.arrows,positioning,calc,spy,intersections}
\pgfplotsset{compat=newest} 
\pgfkeys{/pgf/number format/.cd,1000 sep={}}
\usepackage{placeins}

\newlength\figureheight
\newlength\figurewidth

\pgfplotsset{every axis/.append style={
  grid style={line width=0.6pt,dotted,gray}}}
  
\definecolor{mycolor0}{rgb}{0.2667,0.4471,0.7098}
\definecolor{mycolor1}{rgb}{0.1647,0.6706,0.3804}
\definecolor{mycolor2}{rgb}{0.8275,0.2627,0.3059}
\definecolor{mycolor3}{rgb}{0.5216,0.4392,0.7176}
\definecolor{mycolor4}{rgb}{0.8118,0.7255,0.4118}
\definecolor{mycolor5}{rgb}{0.2745,0.7176,0.8157}

\definecolor{mylcolor0}{rgb}{0.6902,0.7686,0.8863}
\definecolor{mylcolor1}{rgb}{0.5451,0.8902,0.6941}
\definecolor{mylcolor2}{rgb}{0.9412,0.7490,0.7647}
\definecolor{mylcolor3}{rgb}{0.8627,0.8392,0.9176}
\definecolor{mylcolor4}{rgb}{0.9569,0.9373,0.8667}
\definecolor{mylcolor5}{rgb}{0.7529,0.9020,0.9373}
\definecolor{mylcolor6}{rgb}{0.8750,0.8750,0.8750}

\usepackage{bm}
\newcommand{\mathbold}[1]{\bm{#1}}
\newcommand{\mbf}[1]{\mathbf{#1}}

\usepackage{xspace}
\newcommand{\eg}{\textit{e.g.}\xspace}
\newcommand{\ie}{\textit{i.e.}\xspace}
\newcommand{\cf}{\textit{cf.}\xspace}

\newcommand{\T}{\top}
\newcommand{\dd}{\,\mathrm{d}}
\newcommand{\E}{\mathbb{E}}
\newcommand{\R}{\mathbb{R}}
\newcommand{\N}{\mathrm{N}}

\DeclareMathOperator{\diag}{diag}

\newcommand{\GP}{\mathcal{GP}}
\newcommand{\ELBO}{\underbar{$\mathcal{L}$}}
\newcommand{\Z}{\mathcal{Z}}

\newcommand{\vbeta}[0]{\mathbold{\beta}}

\newcommand{\vmu}[0]{\mathbold{\mu}}
\newcommand{\vsigma}[0]{\mathbold{\sigma}}

\renewcommand{\mid}[0]{\,|\,}

\newcommand{\vlambda}[0]{\mathbold{\lambda}}
\newcommand{\vzero}[0]{\mathbold{0}}

\newcommand{\rp}{\mathrm{p}}
\newcommand{\rf}{\mathrm{f}}

\newcommand{\rs}{\mathrm{s}}

\newcommand{\vf}{\mbf{f}}
\newcommand{\vg}{\mbf{g}}
\newcommand{\vh}{\mbf{h}}

\newcommand{\vk}{\mbf{k}}

\newcommand{\vm}{\mbf{m}}

\newcommand{\vp}{\mbf{p}}
\newcommand{\vq}{\mbf{q}}

\newcommand{\vy}{\mbf{y}}

\newcommand{\MA}{\mbf{A}}

\newcommand{\MF}{\mbf{F}}
\newcommand{\MG}{\mbf{G}}

\newcommand{\MK}{\mbf{K}}
\newcommand{\ML}{\mbf{L}}

\newcommand{\MP}{\mbf{P}}
\newcommand{\MQ}{\mbf{Q}}

\newcommand{\MV}{\mbf{V}}

\copyrightnotice{978-1-7281-6662-9/20/\$31.00 {\copyright}2020 IEEE}

\toappear{2020 IEEE International Workshop on Machine Learning for Signal Processing, Sept.\ 21--24, 2020, Espoo, Finland}

\title{Fast Variational Learning in State-Space Gaussian Process Models}

\name{%
    Paul E. Chang$^{\star}$%
    \qquad William J. Wilkinson$^{\star}$%
    \qquad Mohammad Emtiyaz Khan$^{\dagger}$%
    \qquad Arno Solin$^{\star}$%
}
\address{%
    $^{\star}$  Aalto University, Espoo, Finland \\%
    $^{\dagger}$ RIKEN Center for AI Project, Tokyo, Japan%
}

\begin{document}

\maketitle

\begin{abstract}
Gaussian process (GP) regression with 1D inputs can often be performed in linear time via a stochastic differential equation formulation.
However, for non-Gaussian likelihoods, this requires application of approximate inference methods which can make the implementation difficult, \eg, expectation propagation can be numerically unstable and variational inference can be computationally inefficient. In this paper, we propose a new method that removes such difficulties. Building upon an existing method called conjugate-computation variational inference, our approach enables linear-time inference via Kalman recursions while avoiding numerical instabilities and convergence issues. We provide an efficient JAX implementation which exploits just-in-time compilation and allows for fast automatic differentiation through large for-loops. Overall, our approach leads to fast and stable variational inference in state-space GP models that can be scaled to time series with millions of data points.

\end{abstract}
\begin{keywords}
  State-space models, variational inference, Gaussian processes, automatic differentiation
\end{keywords}
\section{Introduction}
\label{sec:intro}
Gaussian process (GP, \cite{rasmussen2006gaussian}) models are non-parametric probabilistic tools shown to be effective in a variety of data analysis tasks. Their main drawback is $O(n^3)$ computational cost of inference, where $n$ is the number of data examples. For a non-Gaussian likelihood, this becomes even more challenging due to the lack of a closed-form expression for the posterior. Developing low-cost algorithms is therefore essential to facilitate application of GPs to real-world problems.

Formulating a GP as a state-space model is one way to reduce the complexity of GP regression to $O(n)$. For one-dimensional inputs, we can do so by using an equivalent stochastic differential equation (SDE) formulation~\cite{Sarkka+Solin:2019} and extend to non-Gaussian problems by using state-of-the-art approximate inference methods~\cite{nickisch2018}, \eg, expectation propagation (EP, \cite{rasmussen2006gaussian}). Doing so allows us to employ Kalman recursions which have $O(n)$ computation and memory cost. Unfortunately, EP can suffer from numerical instability and convergence issues when the model exhibits highly nonlinear behaviour.~Variational inference (VI) is another popular choice which does not have such problems but it often requires $O(n^2)$ memory and cannot be conveniently implemented using Kalman recursions~\cite{opper2009, hensman2015}.~Generally, even when Kalman recursions are used for such approximate inference methods, the practical implementation can be slow since it involves large for-loops, preventing the application of modern automatic differentiation techniques to optimise hyperparameters.~Our goal in this paper is to remove such difficulties and enable fast learning.

\begin{figure}[t!]
  \centering\scriptsize
  \pgfplotsset{yticklabel style={rotate=90}, ylabel style={yshift=0pt},scale only axis,axis on top,clip=true,clip marker paths=true}
  \setlength{\figurewidth}{.42\textwidth}
  \setlength{\figureheight}{.6\figurewidth}
  \input{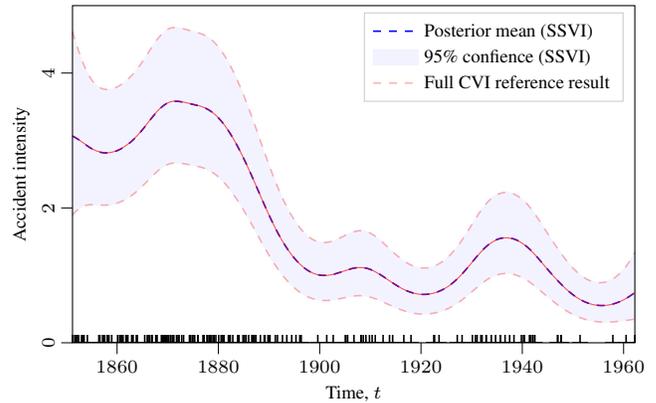}\\[0em]
  \caption{Log-Gaussian Cox process applied to {\em coal mining accidents} data. Our proposed method, SSVI, recovers the same posterior with $O(n)$ computation as the $O(n^3)$ CVI method.}
  \label{fig:teaser}
\end{figure}

We build upon a VI method called conjugate-computation variational inference (CVI, \cite{khan2017}). CVI converts non-Gaussian likelihoods to Gaussian ones, enabling the application of Kalman smoothing to perform inference with $O(n)$ memory and computation cost. To handle large for-loops during hyperparameter learning, we provide an efficient JAX~\cite{jax2018} implementation which employs just-in-time compilation and specifically avoids loop `unrolling'. The resulting method, which we call State-Space VI (SSVI), enables fast learning on data containing more than a million points (see \cref{fig:timings}). The updates of SSVI are identical to CVI (\cref{fig:teaser}), and  strikingly similar to EP, but do not suffer from numerical instability or convergence issues. Comparisons on real-world data demonstrate the efficiency of our method.

\section{Background}
\label{sec:background}
Gaussian processes form a non-parametric family of probability distributions on function spaces, and are completely characterized by a covariance function $\kappa(t,t'): \mathbb{R} \times \mathbb{R} \to \mathbb{R}$ and a mean function which we assume to be zero. Let $\{(t_i, y_i)\}_{i=1}^n$ denote a set of $n$ input--output pairs, then GP models typically take the form
\begin{equation}
f(t) \sim \GP(0, \kappa(t,t') ), \quad \,\,
\vy \mid \vf \sim \prod_{i=1}^{n} p(y_{i} \mid f_i(t_i)), \label{eq:GP-prior-lik}
\end{equation}
which defines the prior for the latent function $f: \R \to \R$ and the likelihood model for $y_i$.
For Gaussian likelihoods, the posterior distribution $p(\vf\mid\vy)$ is Gaussian and can be obtained analytically, but this requires $O(n^3)$ computation in general.
For non-Gaussian likelihoods, the computational overhead is even larger since the posterior is generally intractable and iterative approximate inference must be applied.
Development of efficient, low-cost algorithms for GP models is an important area of research.

Fortunately, as discussed in \cite{Sarkka+Solin:2019}, it is often possible to reformulate GP priors as state-space models which, for the Gaussian likelihoods, reduces the computation cost to $O(n)$. Many widely used covariance functions admit this form exactly or approximately (\eg, the Mat\'ern class, polynomial, noise, constant, squared-exponential, rational quadratic, periodic, and sums/products thereof). The general approach is to rewrite the GP as a linear time-invariant SDE, which has the general continuous-discrete (see \cite{Sarkka+Solin:2019}, p.~200) form:
\begin{align}
\dd\vf(t) &= \MF \, \vf(t)\dd t + \ML \dd \vbeta(t), \label{eq:sde}\\
y_i &\sim p(y_i \mid f_i=\vh^\T \vf(t_i)), \label{eq:sde-lik}
\end{align}
where $\vf(t) \in \R^d$ is the state and $y_i$ is the measurement obtained at time instant $t_i$ via the measurement vector $\vh \in \R^{d}$. $\MF \in \R^{d \times d}$ is the feedback matrix, and $\ML \in\R^{d \times s}$ is the dispersion matrix. $\vbeta(t) \in \R^s$ is the Brownian motion with diffusion matrix $\MQ \in \R^{s \times s} $.
For Gaussian likelihoods, inference then can be performed in $O(n)$ by using Kalman recursions on the above model. This drastic reduction in computation makes the SDE approach an attractive alternative for inference in state-space GP models. 

\subsection{Issues with Learning in State-Space GP Models}
\label{subset:issues}
For non-Gaussian likelihoods, Kalman recursions cannot be applied directly and an approximate inference method is required for tractability. Unfortunately, application of such methods brings new challenges.
Methods such as expectation propagation (EP) provide a Gaussian approximation of the non-Gaussian likelihoods, which can then be used in \cref{eq:sde-lik} to perform inference with Kalman recursions~\cite{wilkinson19}.
Regrettably, EP suffers from numerical issues and is not guaranteed to converge.
Methods such as variational inference (VI) do not have these issues, but standard VI does not provide an EP-like Gaussian approximation of the non-Gaussian likelihood.
Instead, a full-Gaussian approximation over $\vf(t)$ is sought which requires storing a full $n \times n$ covariance matrix during optimization of the variational objective.
It is possible to reparameterize the variational objective to reduce the number of parameters to $O(n)$ \cite{opper2009, nickisch2008}, but still a Kalman recursion algorithm cannot be directly applied to obtain a fast linear-time implementation. 
In general, these two types of methods suffer from different kinds of issues, making inference challenging.

A further issue is in the implementation of such methods. Despite the theoretical guarantee of linear time complexity, Kalman recursions involve large for-loops, making implementation in automatic differentiation frameworks problematic. Previous attempts to overcome this exploit the sparse structure in the precision matrices \cite{durrande2019} or inducing points \cite{adam2020}. This typically requires a full reformulation of the model which can be unstable and difficult to implement. In this paper, we propose a new method that avoids these difficulties.

\section{Methods}
We utilise conjugate-computation variational inference (CVI,  \cite{khan2017}), forming our approximate inference problem in the variational setting by choosing a Gaussian approximate posterior $q(\vf)=\N(\vm,\MV)$ whose natural parameters are $\vlambda^{(1)} {=} \MV^{-1}\vm$ and $\vlambda^{(2)} {=} -\frac{1}{2}\MV^{-1}$. Variational inference aims to optimise $\vlambda$ to maximise a lower bound of  model evidence $p(\vy)$:
\begin{equation}\label{eq:elbo-def}
\log p(\vy)  \geq  \mathbb{E}_{q(\vf)}\bigg[\log \frac{p(\vy, \vf)}{q(\vf)}\bigg] := \ELBO (\vlambda).
\end{equation}
CVI performs natural gradient ascent on the above lower bound. Natural gradients are a way of taking gradient steps that accounts for the informational geometry of the optimisation problem.
CVI utilises the result from \cite{opper2009}, which states that the variational posterior is a sum of the natural parameters of the prior, $\vlambda_{\mathrm{prior}}^{(1)}=\bm{0}$ and  $\vlambda_{\mathrm{prior}}^{(2)} = \frac{1}{2}\MK^{-1}$ and the approximated likelihood terms. CVI finds Gaussian approximations for the non-Gaussian likelihoods with natural parameters $\tilde{\vlambda}=\{\tilde{\vlambda}^{(1)}, \tilde{\vlambda}^{(2)}\}$. The likelihood distribution becomes
\begin{equation}
 p(y_i\mid f_i) \approx \N(\tilde{y}_i \mid f_i, \tilde{\sigma}^2_i),  
\end{equation}
where $\tilde{y}_i = \tilde{\lambda}^{(1)}_i/\tilde{\lambda}^{(2)}_i $ and $\tilde{\sigma}^2_i = -1/(2\,\tilde{\lambda}_i^{(2)})$. Under this parametrisation, $\tilde{\vlambda}$ are the free variational parameters to be optimised.
We can think of performing variational inference in this setting as a series of GP regressions which provide the following approximate posterior: 
\begin{align}\label{eq:post}
  q(\vf) \,{\propto} \bigg[\prod_{i=1}^{n} \N \left(\tilde{y}_i \mid f_{i},\tilde{\sigma}^2_i \right)\!\bigg] \N\left(\vf \mid \vzero, \MK \right).
\end{align}
The likelihood parameters $\tilde{\vlambda}$ are the local variational parameters necessary to compute the global posterior parameters $\vlambda$. We summarise the distributions and parametrisations in the following table.

\smallskip\noindent
\begin{tabularx}{\columnwidth}{c | c | c  }
  \toprule
   Posterior approx & Prior &   Likelihood approx\\
  \midrule
  $\vlambda^{(1)} = \MV^{-1}\vm$ & $\vlambda_{\mathrm{prior}}^{(1)} = \vzero$  & $\tilde{\lambda}^{(1)}_i = \nicefrac{f_i}{\tilde{\sigma}_i^2}$\\ 
  $\vlambda^{(2)} = -\nicefrac{1}{2} \MV^{-1}$ & $\vlambda_{\mathrm{prior}}^{(2)} = -\nicefrac{1}{2} \MK^{-1}$ & $\tilde{\lambda}^{(2)}_i = \nicefrac{-1}{2\tilde{\sigma}^2_i}$ \\
  \bottomrule
\end{tabularx}\medskip

Key to the CVI method is that the natural gradient update can be elegantly computed using the derivatives of the expected log likelihood with respect to the mean parameters $\vmu=\{\vm,\MV+\vm\vm^\T\}$ \cite{khan2017}. 
The two stage update of $\vlambda$ is then:
\begin{align}
\tilde{\vlambda}_{k+1} &= (1-\rho_k)\tilde{\vlambda}_{k} +  \rho_k \, \vg(\vlambda_k), \label{eq:update} \\
\vlambda_{k+1} &= \tilde{\vlambda}_{k+1} + \vlambda_{\mathrm{prior}},  \label{eq:global-update}
\end{align}
where $\vg(\vlambda_k) = \nabla_{\vmu} \E_q \left[\log p(\vy \mid \vf)\right]\,\rvert_{\vmu = \vmu(\vlambda_k)}$, and $\rho_k$ the step size.
Essentially we can update the variational parameters $\tilde{\vlambda}$ using the derivatives of the likelihood terms with respect to parameters of our posterior. Combining local parameter updates and global conjugate regression steps avoid the need to directly optimise \cref{eq:elbo-def}, which alleviates the performance issues with the optimal reduced parametrisation (\cf\ \cref{subset:issues}).

\subsection{Evidence Lower Bound (ELBO) Derivation}
\label{subset:ELBO}
Although the CVI method sidesteps direct computation of the ELBO for the variational parameter updates, it is still required for hyperparameter learning (\eg, kernel length-scale and magnitude). We simplify \cref{eq:elbo-def} by using \cref{eq:post} to give
\begin{equation}
\hspace*{-6pt}\ELBO = \,\E_{q(\vf)} \log  \left[\frac{\prod_{i=1}^{n} p\left(y_{i} \mid f_{i} \right) \Z \left(\GP\right) \N \left(\vf \mid \ \vzero,\MK \right)}
{ \prod_{i=1}^{n}\! \N \big( \tilde{y}_i \mid f_i, \tilde{\sigma}^2_i \big) \N(\vf \mid \vzero, \MK)}\right], \label{eq:elbo-f} 
\end{equation}
where the log marginal likelihood of the approximate conjugate model is
\vspace*{-6pt}
\begin{equation}
\log \Z\left(\GP\right) = -\frac{1}{2}\log|\mathbf{K}_{\tilde{y}}| -\frac{1}{2}\tilde{\vy}^\top\mathbf{K}_{\tilde{y}}^{-1}\tilde{\vy} - \frac{n}{2}\log(2\pi)
\end{equation}
for $\mathbf{K}_{\tilde{y}} := \mathbf{K}+\diag(\tilde{\vsigma}^{2})$. From \cref{eq:elbo-f}, we get the complete expression for the ELBO:
\vspace*{-6pt}
\begin{multline}\label{eq:elbo_simp}
=  \sum_{i=1}^{n} \E_{q(\vf)} \log p\left(y_{i} \mid f_{i}\right) + \log \Z\left(\GP\right)   \\
 - \sum_{i=1}^{n} \bigg[  \frac{1}{2}  \log \bigg(\frac{1}{2 \pi \tilde{\sigma}^2_i}\bigg)
-\frac{1}{2 \tilde{\sigma}^2_i} \bigg( \big(\tilde{y}_i -m_{i}\big)^{2} + v_{i}\bigg)\bigg].
\end{multline}

\subsection{Proposed Method}
\label{sec:methods}
Updating our variational parameters $\vlambda_{k+1}$ in \cref{eq:global-update} involves solving a GP regression problem, which scales as $O(n^3)$. We now show how we can perform the same calculations in $O(n)$ using the Kalman filter and smoother, as well as demonstrating how the CVI updates can be used in the forward filter to initialise the variational parameters. We additionally discuss how the marginal likelihood approximation can be computed as an alternative to the ELBO for hyperparameter learning, and outline our efficient implementation in JAX.

\subsection{Sequential CVI by Filtering and Smoothing} \label{sec:KF-comps}
Certain LTI SDEs of the form in \cref{eq:sde} have discrete-time solutions that can be computed in closed form and written:
\begin{equation}
\vf_{i} = \MA_{i-1}\vf_{i-1} + \vq_{i}, \quad\quad  \vq_{i-1} \sim \N(\vzero,\MQ_{i-1}),
\end{equation}
where $\vf_i=\vf(t_i)$ and $\MA_{i}=\exp(\MF \Delta t_i)$ is the linear state transition matrix, for time step size $\Delta t_i=t_i-t_{i-1}$. $\MQ_{i}$ is the process noise covariance.
For a Gaussian likelihood we can write the measurement model as a linear transformation of state vector with additive Gaussian noise:
\begin{equation}
y_i = \vh^\T\vf_i + \varepsilon_i, \qquad \varepsilon_i \sim \N(0,\sigma^2_i).
\end{equation}
$\vh \in \R^d $ is the measurement vector such that $f(t_i)=\vh^\top \vf_i$, which coincides with the GP model in \cref{eq:GP-prior-lik}. The exact solution to the model outlined above can be computed via the Kalman filter and Rauch--Tung--Striebel smoother (see \cite{Sarkka:2013}). For our non-Gaussian likelihood model, we must adjust the above to reflect our variational likelihood approximations,
\begin{equation}
\tilde{y}_i = \vh^\T\vf_i + \varepsilon_i,  \quad\quad \varepsilon_i \sim \N(0,\tilde{\sigma}^2_i).
\end{equation}
We now derive our adjusted filtering and smoothing algorithms that explicitly incorporate the update steps of \cref{eq:update} and \cref{eq:global-update}, as well as computing all required elements of \cref{eq:elbo_simp}: $\{m_i, \, v_{i}, \, \Z\left(\GP\right), \, \E_{q(\vf)} \log p\left(y_{i} \mid f_{i}\right) \}$.

The filtering distribution, $p(\vf_i \mid \vy_{1:i})=\N(\vf_i \mid \vm_i^{\rf}, \MP_i^{\rf})$, is computed in two stages. Firstly, the prediction step,
\begin{align} \label{eq:prediction}
\vm_{i}^\rp&= \MA_{i}\vm_{i-1}^{\rf}, \qquad \MP_{i}^\rp= \MA_{i} \MP_{i-1}^{\rf} \MA_{i}^\T + \MQ_i,
\end{align}
followed by the update step, in which we first compute the innovation mean ($\eta_i$) and variance ($s_i$),
\begin{equation}
 \eta_i = \tilde{y}_{i} - \vh^\top\vm_{i}^\rp, \qquad s_i = \vh^\top\MP_{i}^\rp\vh + \tilde{\sigma}^2_i.
\end{equation}
The log marginal likelihood of the Gaussian model, as required in \cref{eq:elbo_simp}, can now be evaluated from the above quantities: $\log \mathcal{Z}_i\left(\mathcal{GP}\right) = \sum_{i}^{n} \frac{1}{2} ( \log 2 \pi s_i + \eta_i^2 / s_i)$. The updated filter mean and covariance are then
\begin{align}
\vk_i &= \MP_i^\rp\vh/s_i, \nonumber \label{eq:filter} \\
\vm_{i}^\rf &= \vm_{i}^{\vp} + \vk_i \eta_i, \quad \MP^\rf_i = \MP_i^\rp - \vk_i\vh^\T \MP_i^\rp.
\end{align}
The marginal smoothing distribution is notated $p(\vf_i \mid \vy_{1:n})=\N(\vf_i \mid \vm_i^\rs, \MP_i^\rs)$, and is computed through backward recursion of the following equations:
\begin{align}
\vm_i^\rs &= \vm^\rf_i + \MG_i \left( \vm_{i+1} - \vm_{i+1}^\rp\right), \\
\MP_i^\rs &= \MP^\rf_i + \MG_i(\MP_{i+1} - \MP_{i+1}^\rp)\MG_i^\T,
\end{align}
where $\MG_i {=} \MP^\rf_i \MA^\top_{i+1} [\MP^\rp_{i+1} ]^{-1}$ is the smoother gain. The smoothing distribution gives the GP marginal posterior $q(f(t_i)){=}\N(f(t_i) \mid m_i,v_i)$ where $m_i{=}\vh^\top \vm_i^\rs$ and $v_i{=}\vh^\top \MP_i^\rs \vh$.

\subsection{Variational Parameter Updates in the Filter/Smoother} \label{sec:param_update}
The final required term in order to update $\tilde{\vlambda}$ and compute the ELBO is the variational expectations, \mbox{$\E_{q(\vf)} \log p\left(y_{i} \mid f_{i}\right)$}, and their derivatives. Given we have just computed $m_i$ and $v_i$, it is natural to perform this calculation in the smoother step. The derivatives with respect to the mean parameters $\mu_i=\{m_i,v_i+m_i^2\}$ directly provide the natural parameter update. Using the chain rule, we write down the following update rule at step $k$ as a function of the source parameters $m_i$, $v_i$:
\begin{align}
\mathcal{J}_i &= \E_{\N(f_i \mid m_i, v_i)} \left[ \log p(y_i \mid f_i) \right], \vphantom{\bigg(} \label{eq:smoother_updates}\\
\tilde{\lambda}^{(1)}_{i,k+1} &= (1-\rho_k)\tilde{\lambda}^{(1)}_{i,k} + \rho_k \left(\frac{\partial \mathcal{J}_i}{\partial m_i} - 2\frac{\partial \mathcal{J}_i}{\partial v_i} m_i \right), \label{eq:smoother_updates2}\\
\tilde{\lambda}^{(2)}_{i,k+1} &= (1-\rho_k)\tilde{\lambda}^{(2)}_{i,k} + \rho_k \frac{\partial \mathcal{J}_i}{\partial v_i} . \label{eq:smoother_updates3}
\end{align}
In the general case $\mathcal{J}_i$ is intractable, and we employ Gauss--Hermite quadrature to compute this quantity and its derivatives numerically.
Crucially, these parameter updates are not specific to the smoother, and can also be used in the first forward filtering pass as a novel way to initialise the variational parameters. Initialising the variational distribution to $\N(\bm{0}, \bm{\infty})$ is standard practice, but by letting $m_i=\vh^\top \vm_i^\rf$ and $v_i=\vh^\top \MP_i^\rf \vh$, \ie\ using the marginal \emph{filtering} distribution, and setting $\rho_k=1$, we can utilise \cref{eq:smoother_updates}--\cref{eq:smoother_updates3} to provide a much improved initialisation. In \cref{sec:experiments} we show how doing so results in superior convergence rates in practice.

This interpretation also shows that our CVI scheme can be seen as a new, general purpose nonlinear Kalman filter, whose nonlinear updates equate to a full natural gradient step in the evidence lower bound, and which reduces to the linear Kalman filter when the observation model is Gaussian.

\subsection{Similarity to Expectation Propagation} \label{sec:EP}
It is worth noting that the parameter updates in \cref{sec:param_update} bear striking resemblance to the analogous updates used in filter-smoother version of expectation propagation \cite{wilkinson19}, which can be written with $\mathcal{Z}_i = \log\E_{\N(f_i \mid m_i, v_i)} \left[ p(y_i \mid f_i) \right]$ as
\begin{align} \label{eq:ep_updates}
\hspace*{-8pt}\tilde{\lambda}^{(2)}_{i,\mathrm{new}} &{=} \frac{1}{2}\left( v_i + {\frac{\partial \mathcal{Z}_i}{\partial v_i}}^{-1} \right)^{-1}, \\
\hspace*{-8pt}\tilde{\lambda}^{(1)}_{i,k+1} &{=} (1{-}\rho_k)\tilde{\lambda}^{(1)}_{i,k} {+}  \rho_k  \tilde{\lambda}^{(2)}_{i,\mathrm{new}}\left({2}\frac{\partial \mathcal{Z}_i}{\partial v_i} {\frac{\partial \mathcal{Z}_i}{\partial v_i}}^{-1} {-} {2}m_i \right), \\
\hspace*{-8pt}\tilde{\lambda}^{(2)}_{i,k+1} &{=} (1{-}\rho_k)\tilde{\lambda}^{(2)}_{i,k} + \rho_k \tilde{\lambda}^{(2)}_{i,\mathrm{new}},
\end{align}
where $m_i$, $v_i$ are now the parameters of the so called \emph{cavity} distribution obtained by removing the likelihood from the marginal posterior: $v_i = ((\vh^\top \MP_i^\rs \vh)^{-1} - \tilde{\lambda}^{(2)}_i)^{-1}$ and $m_i = v_i((\vh^\top \MP_i^\rs \vh)^{-1} \vh^\top \vm_i^\rs - \tilde{\lambda}^{(1)}_i)$.

\subsection{Direct Marginal Likelihood Computation} \label{sec:ML}
In sequential models we also have available the marginal likelihood as an alternative to the ELBO as an optimisation objective for hyperparameter learning. The marginal likelihood can be written as a product of conditional terms,
\begin{equation}
  p(\vy) = p(y_1)\,p(y_2 \mid y_1)\,p(y_3 \mid \vy_{1:2})\prod_{i=4}^n p(y_i \mid \vy_{1:i-1}).
\end{equation}
Each term can be computed via numerical integration during the Kalman filter by noticing that,
\begin{align}
\!\!p(y_i \mid \vy_{1:i-1})\, &{=} \int p(y_i \mid \vf_i,\vy_{1:i-1}) p(\vf_i \mid \vy_{1:i-1})\dd\vf_{i} \nonumber \\
 &{=} \int p(y_i \mid f_i = \vh^\T\vf_i) p(\vf_i \mid \vy_{1:i-1})\dd\vf_{i}.
\end{align} 
The first component in the integral is the likelihood, and the second term is the filter prediction calculated in \cref{eq:prediction}.

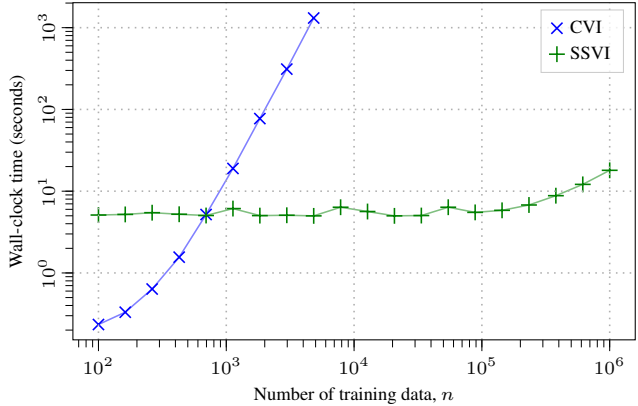
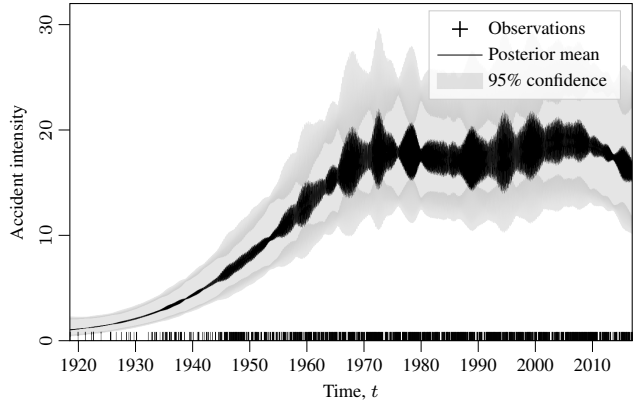
\begin{figure*}[!t]
  \centering\scriptsize
  \pgfplotsset{yticklabel style={rotate=90}, ylabel style={yshift=0pt},scale only axis,axis on top,clip=true,clip marker paths=true}
  \setlength{\figurewidth}{.42\textwidth}
  \setlength{\figureheight}{.6\figurewidth}
  \begin{subfigure}[b]{.48\textwidth}
    \centering
\begin{tikzpicture}

\begin{axis}[
height=\figureheight,
legend cell align={left},
legend style={fill opacity=0.8, draw opacity=1, text opacity=1, draw=white!80!black},
log basis x={10},
log basis y={10},
tick align=outside,
tick pos=left,
width=\figurewidth,
x grid style={white!69.0196078431373!black},
xlabel={Number of training data, \(\displaystyle n\)},
xmajorgrids,
xmin=63.0957344480193, xmax=1584893.19246111,
xmode=log,
xtick style={color=black},
y grid style={white!69.0196078431373!black},
ylabel={Wall-clock time (seconds)},
ymajorgrids,
ymin=0.151984175577783, ymax=2020.60808640779,
ymode=log,
ytick style={color=black}
]
\addplot [semithick, blue, opacity=0.5, forget plot]
table {%
100 0.234012222290039
162 0.330991840362549
263 0.636063241958618
428 1.56023635864258
695 5.1824866771698
1128 18.9748750209808
1832 77.2779819488525
2976 311.475989437103
4832 1312.32655787468
7847 nan
};
\addplot [semithick, green!50!black, opacity=0.5, forget plot]
table {%
100 5.10986604690552
162 5.19853620529175
263 5.45848460197449
428 5.22295546531677
695 5.03991341590881
1128 6.13004679679871
1832 5.02756352424622
2976 5.07760033607483
4832 4.99036955833435
7847 6.36353397369385
12742 5.6413978099823
20691 4.99379358291626
33598 5.03634300231934
54555 6.36611557006836
88586 5.50497102737427
143844 5.83312559127808
233572 6.78001189231873
379269 8.81223740577698
615848 12.1159058094025
1000000 18.0030683994293
};
\addplot [semithick, blue, opacity=1.0, mark=x, mark size=3, mark options={solid}, only marks]
table {%
100 0.234012222290039
162 0.330991840362549
263 0.636063241958618
428 1.56023635864258
695 5.1824866771698
1128 18.9748750209808
1832 77.2779819488525
2976 311.475989437103
4832 1312.32655787468
7847 nan
12742 nan
20691 nan
33598 nan
54555 nan
88586 nan
143844 nan
233572 nan
379269 nan
615848 nan
1000000 nan
};
\addlegendentry{CVI}
\addplot [semithick, green!50!black, opacity=1.0, mark=+, mark size=3, mark options={solid}, only marks]
table {%
100 5.10986604690552
162 5.19853620529175
263 5.45848460197449
428 5.22295546531677
695 5.03991341590881
1128 6.13004679679871
1832 5.02756352424622
2976 5.07760033607483
4832 4.99036955833435
7847 6.36353397369385
12742 5.6413978099823
20691 4.99379358291626
33598 5.03634300231934
54555 6.36611557006836
88586 5.50497102737427
143844 5.83312559127808
233572 6.78001189231873
379269 8.81223740577698
615848 12.1159058094025
1000000 18.0030683994293
};
\addlegendentry{SSVI}
\end{axis}

\end{tikzpicture}\\[0em]
    \caption{Number of data vs.\ wall-clock time}
 	\label{fig:timings}
  \end{subfigure}
  \hfill
  \begin{subfigure}[b]{.48\textwidth}
    \centering
    \pgfplotsset{scaled x ticks = false}
\begin{tikzpicture}

\begin{axis}[
height=\figureheight,
legend cell align={left},
legend style={fill opacity=0.8, draw opacity=1, text opacity=1, draw=white!80!black},
tick align=outside,
tick pos=left,
width=\figurewidth,
x grid style={white!69.0196078431373!black},
xlabel={Time, \(\displaystyle t\)},
xmin=701101, xmax=737060,
xtick style={color=black},
xtick={701647,705299.5,708952,712604.5,716257,719909.5,723562,727214.5,730867,734519.5},
xticklabels={1920,1930,1940,1950,1960,1970,1980,1990,2000,2010},
y grid style={white!69.0196078431373!black},
ylabel={Accident intensity},
ymin=0, ymax=32,
ytick style={color=black}
]
\addplot [semithick, black, mark=+, mark size=3, mark options={solid}, only marks]
table {%
0 -3
};
\addlegendentry{Observations}
\addplot [line width=0.08pt, black]
table {%
0 -3
};
\addlegendentry{Posterior mean}
\path [draw=black, fill=black, opacity=0.1]
(axis cs:0,-3)
--(axis cs:0,-1)
--(axis cs:1,-1)
--(axis cs:1,-3)
--(axis cs:1,-3)
--(axis cs:0,-3)
--cycle;
\addlegendimage{area legend, draw=black, fill=black, opacity=0.1}
\addlegendentry{95\% confidence}

\addplot graphics [includegraphics cmd=\pgfimage,xmin=695301.161290323, xmax=741699.870967742, ymin=-4.57142857142857, ymax=36.987012987013] {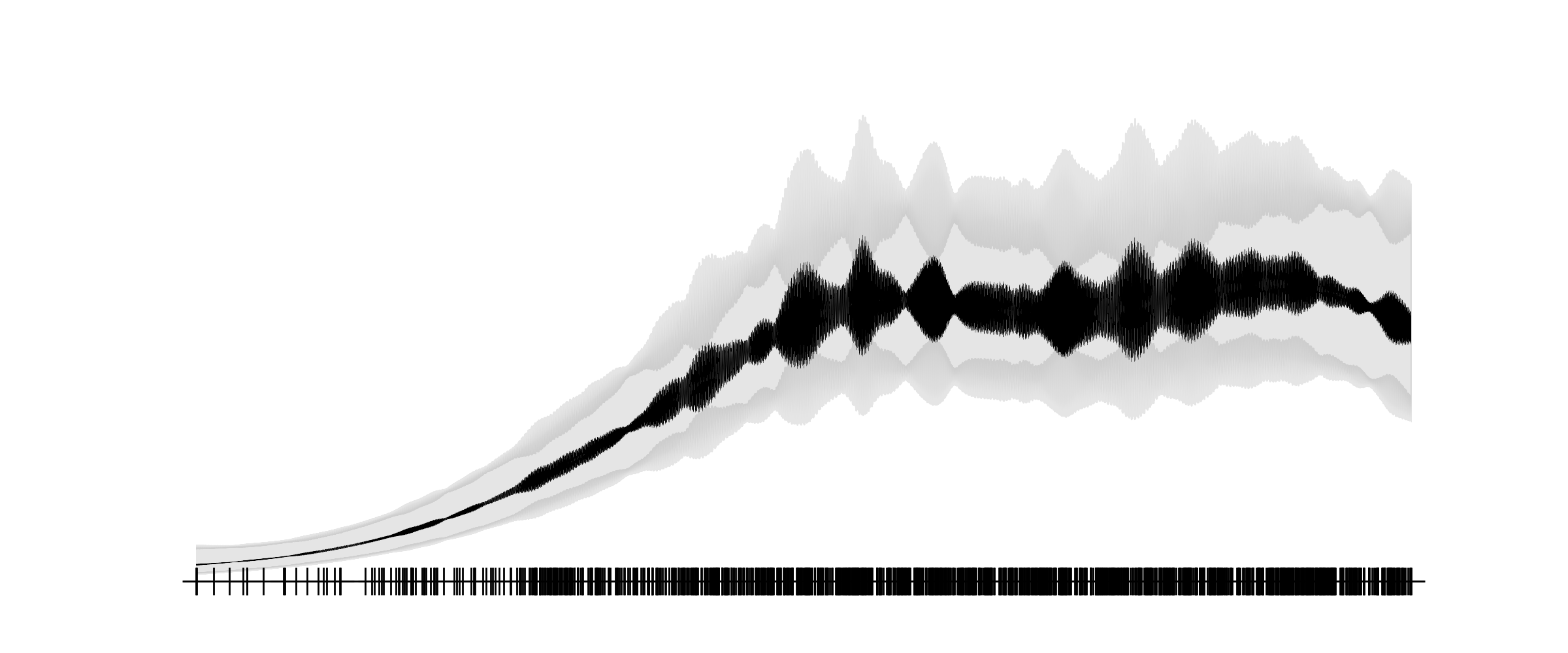};
\end{axis}

\end{tikzpicture}\\[0em]
    \caption{Airline accidents ($n \approx 40\,\textrm{k}$)}
    \label{fig:airline}
  \end{subfigure}
  \vspace*{-.5em}
  \caption{(a)~Empirical wall-clock times (averages over 5~runs) for taking ten variational steps and evaluating the ELBO and its gradient for hyperparameter learning in a simulated GP classification task. CVI is implemented in GPflow~2 and shows cubic computational cost in $n$. SSVI delivers the same result in linear time. The set-up cost (${\sim}6$~s) in SSVI is dominating up to $10^5$, whereafter the scaling is linear (subsequent iteration steps do not suffer from the setup time). (b)~The posterior for the {\em airline accidents} intensity modelling task with a slow trend and multiple periodic kernels for day-of-week and time-of-year effects.}

  \label{fig:comparisons}
\end{figure*}

\subsection{Efficient Hyperparameter Learning with JAX} \label{sec:JAX}
The sequential formulation of GP models is an extremely efficient approach to inference. However, a problem arises in the machine learning context, where it is desirable to optimise the model hyperparameters via gradient-based methods using automatic differentiation. Most automatic differentiation libraries work by `tracing' a computational graph. This involves passing arbitrary values through the supplied functions and constructing a list of the necessary operations and their derivatives. Functions involving large for-loops (such as a Kalman filter) result in massive computational graphs that involve large compilation overheads, memory usage and runtime. For this reason, most machine learning approaches to temporal GPs either use finite differences \cite{nickisch2018}, which are slow when the number of parameters is large, or reformulate the model entirely to exploit linear algebra tricks applicable to sparse precision matrices \cite{durrande2019}.

We utilise the following novel capabilities of the increasingly popular differential programming Python framework, JAX \cite{jax2018}: \emph{(i)}~we avoid `unrolling' of for-loops, \ie\ instead of building a large graph of repeated operations, a smaller graph is recursively called, reducing the compilation overhead and memory. \emph{(ii)}~we just-in-time (JIT) compile the for-loops, to avoid the cost of graph retracing. This results in an overhead setup cost on the first function call (this effect is seen in \cref{fig:timings}), but means that every subsequent call only involves reuse of the static graph, which is very efficient. \emph{(iii)}~JAX also allows for the use of accelerated linear algebra (XLA) to speed up the underlying filtering/smoothing operations. Combined, the above implementation details result in an extremely fast method that scales to millions of data points. \cref{fig:timings} shows that one training iteration for a one-dimensional GP with one million data points takes approximately 20~s.
\section{Experiments}
\label{sec:experiments}
Initial experiments show that SSVI is an efficient inference method for fitting large non-Gaussian time series models. We show that the SSVI posterior is equal to CVI one, as expected. Furthermore, on large datasets our method performs comparably to EP in terms of test performance and convergence speed. In addition to choosing the ELBO as a training objective we also use the marginal likelihood and compare their performance. The practical computational complexity is also shown, by running a wall-clock test on a simple GP classification task. All experiments  were performed using a MacBook pro with a 2.4~GHz Intel core i5 processor and 16~Gb RAM. 
\subsection{Comparison to Full-CVI}
\cref{fig:timings} shows the computation times for CVI (implemented in GPflow~2) vs.\ SSVI on a GP classification example where both return the same solution. The data were simulated from $y_i \sim \text{Bern}(f_i)$, where $f(t) = 6\sin(\frac{\pi t}{10})/\pi \frac{t}{10} + 1$ and a Mat\'ern-$\nicefrac{5}{2}$ GP prior was used. The number of observations $n$ was varied from 100 data points to one million. The chart shows the linear computational complexity in $n$ of SSVI versus the cubic complexity for CVI, noting that SSVI's setup cost dominates until around $n=1,000$. It should be noted that subsequent iterations do not include the setup cost, making optimisation very fast.

The experiment in \cref{fig:teaser} uses the coal mining disaster dataset \cite{wilkinson19} that contains dates for 191 explosions that killed ten or more men in Britain between 1851--1962. We use a log-Gaussian Cox process, which is an inhomogeneous Poisson process (approximated with a Poisson likelihood for 200 equal-time interval bins). We use a Mat\'ern-$\nicefrac{5}{2}$ GP prior with likelihood model $ p(\vy \mid \vf) \approx \prod_{i=1}^n \mathrm{Poisson}(y_i \mid \exp(f(\hat{t}_i)))$, where $\hat{t}_i$ is the bin coordinate and $y_i$ the number of disasters in the bin. Given a small data size we can use CVI and SSVI and compare the posterior mean and variances for both methods after training for 500 iterations using the Adam optimizer. The plot shows negligible difference between the methods.

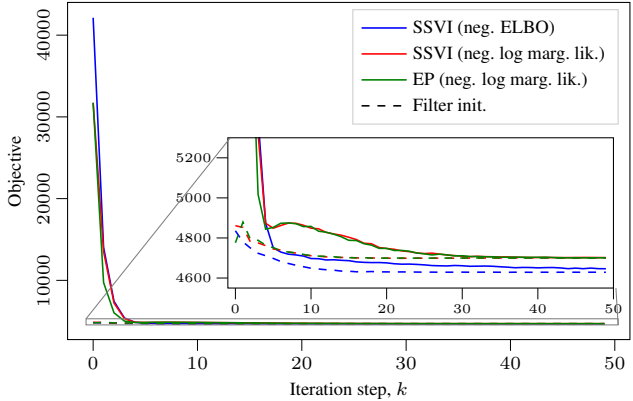
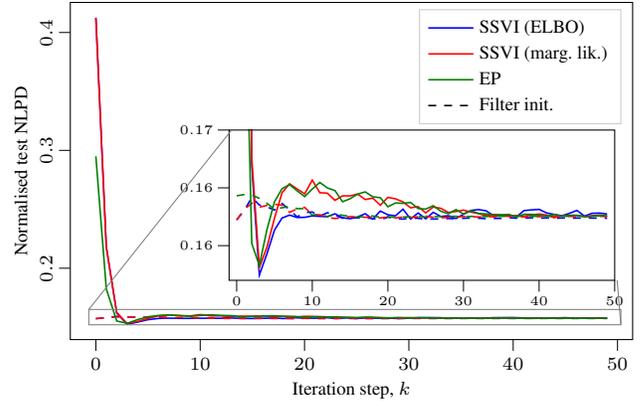
\begin{figure*}[!t]
  \centering\scriptsize
  \tikzstyle{every picture}+=[remember picture]
  \pgfplotsset{yticklabel style={rotate=90}, ylabel style={yshift=0pt},scale only axis,axis on top,clip=true,clip marker paths=true, scaled y ticks = false}
  \setlength{\figurewidth}{.42\textwidth}
  \setlength{\figureheight}{.6\figurewidth}
  \begin{subfigure}[b]{.48\textwidth}
    \centering
\begin{tikzpicture}

\begin{axis}[
height=\figureheight,
legend cell align={left},
legend style={fill opacity=0.8, draw opacity=1, text opacity=1, draw=white!80!black},
tick align=outside,
tick pos=left,
width=\figurewidth,
x grid style={white!69.0196078431373!black},
xlabel={Iteration step, \(\displaystyle k\)},
xmin=-2.45, xmax=51.45,
xtick style={color=black},
y grid style={white!69.0196078431373!black},
ylabel={Objective},
ymin=2753.65359754986, ymax=44003.2323568536,
ytick style={color=black}
]
\addplot [semithick, blue]
table {%
0 42128.251504158
1 14055.2060483232
2 7433.30652020951
3 5398.27899969677
4 4872.49087738629
5 4751.31872432939
6 4729.53337730188
7 4719.04474598364
8 4715.23885351152
9 4709.68159419692
10 4697.26693373311
11 4696.41460976885
12 4690.12949467287
13 4691.15542803183
14 4688.41120752369
15 4686.64246419781
16 4680.64796138832
17 4678.19067540079
18 4677.11142239412
19 4677.30806274131
20 4675.71219320999
21 4671.22288526859
22 4669.0990481442
23 4669.57591357609
24 4667.91765423337
25 4665.62393148804
26 4662.22580078173
27 4661.09613258231
28 4660.35637056232
29 4662.66370919477
30 4661.31461005409
31 4658.61212907679
32 4659.15567687356
33 4659.85027168027
34 4658.7324721022
35 4656.30021447139
36 4654.17335885397
37 4651.34840683082
38 4652.87332266544
39 4652.46717990893
40 4650.79149850772
41 4652.00606774526
42 4652.39306179219
43 4650.38769893646
44 4646.72274274834
45 4649.32222647951
46 4646.61380182512
47 4649.62460942488
48 4646.72637743971
49 4646.341730991
};
\addlegendentry{SSVI (neg. ELBO)}
\addplot [semithick, blue, dashed, forget plot]
table {%
0 4835.54644387874
1 4780.32675489223
2 4749.22441439967
3 4723.45284279966
4 4711.39567554439
5 4698.01048411097
6 4682.50046268167
7 4672.71843090158
8 4663.84420200921
9 4656.4650391371
10 4648.67081616143
11 4644.82685773722
12 4641.37118828511
13 4639.29018317512
14 4634.6664124487
15 4634.00334886455
16 4631.23743472852
17 4631.19608308993
18 4632.39166112368
19 4630.83982149899
20 4630.54793123234
21 4630.28594744782
22 4630.51568663233
23 4630.20278260994
24 4629.77041421308
25 4629.20899910135
26 4629.43198214255
27 4629.06441482043
28 4629.09420691511
29 4629.25456903314
30 4629.401075111
31 4629.04801899305
32 4629.20156171312
33 4629.32177236117
34 4629.27689061257
35 4628.90962437042
36 4628.63445024548
37 4629.15293892505
38 4628.8865265778
39 4628.69767162723
40 4628.92837675595
41 4629.0139468664
42 4628.83179282545
43 4628.96239155237
44 4629.02515248836
45 4628.9880200628
46 4628.72577946481
47 4629.02564267038
48 4628.89048146125
49 4628.70095685449
};
\addplot [semithick, red]
table {%
0 31728.5200852063
1 13530.4005281519
2 7231.06821992504
3 5319.15891095655
4 4873.12922745792
5 4848.71100108785
6 4862.07203151913
7 4874.05113155056
8 4873.02627253566
9 4863.49145421489
10 4848.79903566834
11 4845.6689570561
12 4827.88462942102
13 4820.09452290737
14 4813.63293644956
15 4803.47420811632
16 4792.02540945624
17 4773.65038576284
18 4771.0414817238
19 4752.49010622939
20 4747.4453531861
21 4743.45352048458
22 4735.35790713359
23 4731.57630773093
24 4725.12367865258
25 4722.1171809229
26 4724.29320576523
27 4719.26335536316
28 4714.88992416549
29 4711.96274461138
30 4708.73461964597
31 4706.43475867901
32 4708.59237013061
33 4706.63532571838
34 4706.22538456775
35 4705.27560365081
36 4705.46073076095
37 4704.31078737493
38 4703.4469204314
39 4702.38345219398
40 4702.63830639865
41 4703.36415509028
42 4702.25327999977
43 4701.9463219852
44 4702.12121846294
45 4702.32204120872
46 4702.67244747917
47 4702.56994944211
48 4702.19596729347
49 4702.03514130992
};
\addlegendentry{SSVI (neg.\ log marg.\ lik.)}
\addplot [semithick, red, dashed, forget plot]
table {%
0 4861.25830498643
1 4852.37193243669
2 4778.81303866349
3 4775.05394538196
4 4761.54236175066
5 4744.98442995332
6 4739.90798098314
7 4729.63131243683
8 4724.62914934418
9 4715.65754159168
10 4711.40712716248
11 4708.77926156512
12 4705.68656292194
13 4703.7171491379
14 4702.71063933849
15 4703.22837475759
16 4700.09030072592
17 4701.47094418432
18 4700.19408426816
19 4699.82964133995
20 4698.93284268501
21 4698.67923748671
22 4698.29390465777
23 4699.40056022294
24 4699.31902328791
25 4699.12749725817
26 4699.01689277689
27 4699.454422415
28 4699.21031707889
29 4699.04854222442
30 4699.2481357179
31 4699.53835370807
32 4699.73348927756
33 4699.60384258426
34 4699.5135605849
35 4699.55000966151
36 4699.72578866872
37 4699.86874952109
38 4700.32983131039
39 4700.3065560184
40 4700.16895367637
41 4700.18508620404
42 4700.29265105681
43 4700.55794423791
44 4700.27654775604
45 4700.195107203
46 4700.46227513262
47 4700.29062404127
48 4700.80201781313
49 4700.50613940939
};
\addplot [semithick, green!50!black]
table {%
0 31728.5200852063
1 9734.38483123022
2 6026.94059043575
3 5017.43532926914
4 4843.72100920281
5 4852.64998440182
6 4873.33058853788
7 4874.92208465927
8 4871.06717744626
9 4856.57273987071
10 4857.69892727302
11 4833.61704493979
12 4828.69028515782
13 4818.82022251581
14 4805.8636909087
15 4788.25814426803
16 4787.12479810815
17 4771.23204406135
18 4764.12990278282
19 4748.15094631331
20 4747.93724116676
21 4737.46433343498
22 4737.81865462136
23 4733.76328994415
24 4724.76752719756
25 4720.58652435569
26 4715.23819803739
27 4719.11351652429
28 4715.64143843507
29 4710.45028209892
30 4711.24093992278
31 4707.53200590541
32 4707.28122139541
33 4706.25075380007
34 4705.87504063668
35 4703.88599052584
36 4703.69820697966
37 4704.24529447653
38 4703.45486238836
39 4703.21002568809
40 4703.34719211213
41 4703.69759381123
42 4702.16460472773
43 4701.50538684501
44 4700.41448965535
45 4701.37357530572
46 4700.43822639861
47 4700.95787924719
48 4700.97881088785
49 4701.00355225994
};
\addlegendentry{EP (neg. log marg.\ lik.)}
\addplot [semithick, green!50!black, dashed, forget plot]
table {%
0 4776.18135009352
1 4879.01638421326
2 4791.24062846554
3 4788.19970210134
4 4764.09913885219
5 4751.39655733444
6 4737.09626597291
7 4729.63291711401
8 4725.60624635543
9 4717.7827461474
10 4711.38207150135
11 4709.42060973637
12 4707.91385551439
13 4706.07931448251
14 4702.73156310311
15 4701.37822675811
16 4701.02181532311
17 4701.30849466116
18 4699.27225995876
19 4699.82691030108
20 4699.06413113693
21 4699.73215760489
22 4699.68833875855
23 4699.25954847511
24 4699.29618032724
25 4699.75894650045
26 4699.39777386437
27 4699.55246218923
28 4699.52463056381
29 4699.4540631204
30 4699.91561708038
31 4699.89000590021
32 4700.13534911882
33 4699.76314103048
34 4700.26006407649
35 4700.34447337382
36 4699.97095877854
37 4699.78941255895
38 4700.07957469501
39 4700.29193835105
40 4700.19989845498
41 4700.10758581685
42 4700.27146477665
43 4700.39249325369
44 4700.61434302995
45 4700.49411839264
46 4700.37582128223
47 4700.34193836697
48 4700.52190299388
49 4700.46251938458
};
\addplot [semithick, black, dashed]
table {%
1 2752.65359754986
};
\addlegendentry{Filter init.}
\draw (axis cs:29,16503.5131839844) node[
  scale=0.5,
  anchor=base west,
  text=black,
  rotate=0.0
]{\tikz\node[coordinate] (n1) {};};
\draw (axis cs:-1,5300) node[
  scale=0.5,
  anchor=base west,
  text=black,
  rotate=0.0
]{\tikz\node[coordinate] (r1) {};};
\draw (axis cs:50,4550) node[
  scale=0.5,
  anchor=base west,
  text=black,
  rotate=0.0
]{\tikz\node[coordinate] (r2) {};};
\end{axis}

\end{tikzpicture}\\[0em]
    \caption{Training objective}
    \label{fig:training}

    \begin{tikzpicture}[remember picture,overlay]
       \tiny
       \pgfplotsset{yticklabel style={rotate=-90}}
       \setlength{\figurewidth}{.6\textwidth}
       \setlength{\figureheight}{.39\figurewidth}
       \fill[draw=black!50,fill=none] (r1) rectangle (r2);
       \node at (n1) {
\begin{tikzpicture}

\begin{axis}[
height=\figureheight,
tick align=outside,
tick pos=left,
width=\figurewidth,
x grid style={white!69.0196078431373!black},
xmin=-1, xmax=50,
xtick style={color=black},
y grid style={white!69.0196078431373!black},
ymin=4550, ymax=5300,
ytick style={color=black}
]
\addplot [semithick, blue]
table {%
0 42128.251504158
1 14055.2060483232
2 7433.30652020951
3 5398.27899969677
4 4872.49087738629
5 4751.31872432939
6 4729.53337730188
7 4719.04474598364
8 4715.23885351152
9 4709.68159419692
10 4697.26693373311
11 4696.41460976885
12 4690.12949467287
13 4691.15542803183
14 4688.41120752369
15 4686.64246419781
16 4680.64796138832
17 4678.19067540079
18 4677.11142239412
19 4677.30806274131
20 4675.71219320999
21 4671.22288526859
22 4669.0990481442
23 4669.57591357609
24 4667.91765423337
25 4665.62393148804
26 4662.22580078173
27 4661.09613258231
28 4660.35637056232
29 4662.66370919477
30 4661.31461005409
31 4658.61212907679
32 4659.15567687356
33 4659.85027168027
34 4658.7324721022
35 4656.30021447139
36 4654.17335885397
37 4651.34840683082
38 4652.87332266544
39 4652.46717990893
40 4650.79149850772
41 4652.00606774526
42 4652.39306179219
43 4650.38769893646
44 4646.72274274834
45 4649.32222647951
46 4646.61380182512
47 4649.62460942488
48 4646.72637743971
49 4646.341730991
};
\addplot [semithick, blue, dashed]
table {%
0 4835.54644387874
1 4780.32675489223
2 4749.22441439967
3 4723.45284279966
4 4711.39567554439
5 4698.01048411097
6 4682.50046268167
7 4672.71843090158
8 4663.84420200921
9 4656.4650391371
10 4648.67081616143
11 4644.82685773722
12 4641.37118828511
13 4639.29018317512
14 4634.6664124487
15 4634.00334886455
16 4631.23743472852
17 4631.19608308993
18 4632.39166112368
19 4630.83982149899
20 4630.54793123234
21 4630.28594744782
22 4630.51568663233
23 4630.20278260994
24 4629.77041421308
25 4629.20899910135
26 4629.43198214255
27 4629.06441482043
28 4629.09420691511
29 4629.25456903314
30 4629.401075111
31 4629.04801899305
32 4629.20156171312
33 4629.32177236117
34 4629.27689061257
35 4628.90962437042
36 4628.63445024548
37 4629.15293892505
38 4628.8865265778
39 4628.69767162723
40 4628.92837675595
41 4629.0139468664
42 4628.83179282545
43 4628.96239155237
44 4629.02515248836
45 4628.9880200628
46 4628.72577946481
47 4629.02564267038
48 4628.89048146125
49 4628.70095685449
};
\addplot [semithick, red]
table {%
0 31728.5200852063
1 13530.4005281519
2 7231.06821992504
3 5319.15891095655
4 4873.12922745792
5 4848.71100108785
6 4862.07203151913
7 4874.05113155056
8 4873.02627253566
9 4863.49145421489
10 4848.79903566834
11 4845.6689570561
12 4827.88462942102
13 4820.09452290737
14 4813.63293644956
15 4803.47420811632
16 4792.02540945624
17 4773.65038576284
18 4771.0414817238
19 4752.49010622939
20 4747.4453531861
21 4743.45352048458
22 4735.35790713359
23 4731.57630773093
24 4725.12367865258
25 4722.1171809229
26 4724.29320576523
27 4719.26335536316
28 4714.88992416549
29 4711.96274461138
30 4708.73461964597
31 4706.43475867901
32 4708.59237013061
33 4706.63532571838
34 4706.22538456775
35 4705.27560365081
36 4705.46073076095
37 4704.31078737493
38 4703.4469204314
39 4702.38345219398
40 4702.63830639865
41 4703.36415509028
42 4702.25327999977
43 4701.9463219852
44 4702.12121846294
45 4702.32204120872
46 4702.67244747917
47 4702.56994944211
48 4702.19596729347
49 4702.03514130992
};
\addplot [semithick, red, dashed]
table {%
0 4861.25830498643
1 4852.37193243669
2 4778.81303866349
3 4775.05394538196
4 4761.54236175066
5 4744.98442995332
6 4739.90798098314
7 4729.63131243683
8 4724.62914934418
9 4715.65754159168
10 4711.40712716248
11 4708.77926156512
12 4705.68656292194
13 4703.7171491379
14 4702.71063933849
15 4703.22837475759
16 4700.09030072592
17 4701.47094418432
18 4700.19408426816
19 4699.82964133995
20 4698.93284268501
21 4698.67923748671
22 4698.29390465777
23 4699.40056022294
24 4699.31902328791
25 4699.12749725817
26 4699.01689277689
27 4699.454422415
28 4699.21031707889
29 4699.04854222442
30 4699.2481357179
31 4699.53835370807
32 4699.73348927756
33 4699.60384258426
34 4699.5135605849
35 4699.55000966151
36 4699.72578866872
37 4699.86874952109
38 4700.32983131039
39 4700.3065560184
40 4700.16895367637
41 4700.18508620404
42 4700.29265105681
43 4700.55794423791
44 4700.27654775604
45 4700.195107203
46 4700.46227513262
47 4700.29062404127
48 4700.80201781313
49 4700.50613940939
};
\addplot [semithick, green!50!black]
table {%
0 31728.5200852063
1 9734.38483123022
2 6026.94059043575
3 5017.43532926914
4 4843.72100920281
5 4852.64998440182
6 4873.33058853788
7 4874.92208465927
8 4871.06717744626
9 4856.57273987071
10 4857.69892727302
11 4833.61704493979
12 4828.69028515782
13 4818.82022251581
14 4805.8636909087
15 4788.25814426803
16 4787.12479810815
17 4771.23204406135
18 4764.12990278282
19 4748.15094631331
20 4747.93724116676
21 4737.46433343498
22 4737.81865462136
23 4733.76328994415
24 4724.76752719756
25 4720.58652435569
26 4715.23819803739
27 4719.11351652429
28 4715.64143843507
29 4710.45028209892
30 4711.24093992278
31 4707.53200590541
32 4707.28122139541
33 4706.25075380007
34 4705.87504063668
35 4703.88599052584
36 4703.69820697966
37 4704.24529447653
38 4703.45486238836
39 4703.21002568809
40 4703.34719211213
41 4703.69759381123
42 4702.16460472773
43 4701.50538684501
44 4700.41448965535
45 4701.37357530572
46 4700.43822639861
47 4700.95787924719
48 4700.97881088785
49 4701.00355225994
};
\addplot [semithick, green!50!black, dashed]
table {%
0 4776.18135009352
1 4879.01638421326
2 4791.24062846554
3 4788.19970210134
4 4764.09913885219
5 4751.39655733444
6 4737.09626597291
7 4729.63291711401
8 4725.60624635543
9 4717.7827461474
10 4711.38207150135
11 4709.42060973637
12 4707.91385551439
13 4706.07931448251
14 4702.73156310311
15 4701.37822675811
16 4701.02181532311
17 4701.30849466116
18 4699.27225995876
19 4699.82691030108
20 4699.06413113693
21 4699.73215760489
22 4699.68833875855
23 4699.25954847511
24 4699.29618032724
25 4699.75894650045
26 4699.39777386437
27 4699.55246218923
28 4699.52463056381
29 4699.4540631204
30 4699.91561708038
31 4699.89000590021
32 4700.13534911882
33 4699.76314103048
34 4700.26006407649
35 4700.34447337382
36 4699.97095877854
37 4699.78941255895
38 4700.07957469501
39 4700.29193835105
40 4700.19989845498
41 4700.10758581685
42 4700.27146477665
43 4700.39249325369
44 4700.61434302995
45 4700.49411839264
46 4700.37582128223
47 4700.34193836697
48 4700.52190299388
49 4700.46251938458
};
\draw (axis cs:-1,5300) node[
  scale=0.5,
  anchor=base west,
  text=black,
  rotate=0.0
]{\tikz\node[coordinate] (z1) {};};
\draw (axis cs:50,4550) node[
  scale=0.5,
  anchor=base west,
  text=black,
  rotate=0.0
]{\tikz\node[coordinate] (z2) {};};
\end{axis}

\end{tikzpicture}};
       \draw[draw=black!50] (r1)--(z1);
       \draw[draw=black!50] (r2)--(z2);
	\end{tikzpicture}    
  \end{subfigure}
  \hfill
  \begin{subfigure}[b]{.48\textwidth}
    \centering
\begin{tikzpicture}

\begin{axis}[
height=\figureheight,
legend cell align={left},
legend style={fill opacity=0.8, draw opacity=1, text opacity=1, draw=white!80!black},
tick align=outside,
tick pos=left,
width=\figurewidth,
x grid style={white!69.0196078431373!black},
xlabel={Iteration step, \(\displaystyle k\)},
xmin=-2.45, xmax=51.45,
xtick style={color=black},
y grid style={white!69.0196078431373!black},
ylabel={Normalised test NLPD},
ymin=0.139454416646159, ymax=0.42551065989061,
ytick style={color=black}
]
\addplot [semithick, blue]
table {%
0 0.41250810337944
1 0.218024495284717
2 0.1619033682228
3 0.15245697315727
4 0.154281702135226
5 0.156341195644042
6 0.15738839443198
7 0.157643572121716
8 0.157437142487707
9 0.157448874019766
10 0.15757463808453
11 0.157531814750928
12 0.157507029295473
13 0.157752224758334
14 0.157416203021393
15 0.157328205608468
16 0.157482831869204
17 0.157761949558959
18 0.157295073232787
19 0.157991481387726
20 0.157670139793429
21 0.157575397413035
22 0.158155560504331
23 0.157302788450984
24 0.157826355247713
25 0.157321520157085
26 0.157883027529256
27 0.15752230166456
28 0.157588792400429
29 0.157344748183533
30 0.157720086663781
31 0.158050142661836
32 0.158120381068668
33 0.158010566017018
34 0.157582128123029
35 0.157573668826117
36 0.157655396088931
37 0.157480827273648
38 0.15803137998468
39 0.158091333949074
40 0.158137859385754
41 0.157824940936067
42 0.157703577012098
43 0.15758267475298
44 0.157789794076435
45 0.158046247410907
46 0.157556095578216
47 0.157797009991546
48 0.157817607233681
49 0.157737639643816
};
\addlegendentry{SSVI (ELBO)}
\addplot [semithick, blue, dashed, forget plot]
table {%
0 0.157229059997197
1 0.15803865241808
2 0.159173727487782
3 0.158510630000205
4 0.158362565955827
5 0.158072805437503
6 0.158648140002203
7 0.158092339890234
8 0.157491630668904
9 0.157861849438426
10 0.157910224517489
11 0.1575656668202
12 0.157847163209446
13 0.157640781662066
14 0.157376133292262
15 0.157406705801527
16 0.157391055037299
17 0.1573965683197
18 0.157347548372605
19 0.157232308209555
20 0.157462204893046
21 0.157355148515592
22 0.157296110882323
23 0.157396036239828
24 0.157365436441189
25 0.157292697210476
26 0.157305140168827
27 0.157313431682967
28 0.157345123018259
29 0.157325201497627
30 0.157364202965764
31 0.157397023455578
32 0.157372240751134
33 0.157333461159887
34 0.15735455700473
35 0.157359912635817
36 0.157418187718972
37 0.157379257461651
38 0.157382351677353
39 0.157408855848361
40 0.157365951313816
41 0.157407604212913
42 0.157355155857192
43 0.157399017952772
44 0.157374837735437
45 0.157386403770081
46 0.157386035817008
47 0.157384228925716
48 0.157377506897062
49 0.15737876017368
};
\addplot [semithick, red]
table {%
0 0.412508103379498
1 0.218165079029874
2 0.162520995452032
3 0.153105488735592
4 0.154854441909495
5 0.157364872981406
6 0.159569722440883
7 0.160348722059811
8 0.159901840583811
9 0.159481506671422
10 0.160670282075018
11 0.159577746437354
12 0.159539250959865
13 0.158914380344364
14 0.159304445737739
15 0.159303730678795
16 0.15946046598674
17 0.158988457568873
18 0.159111187006929
19 0.158495003232138
20 0.159177889212118
21 0.159051390328421
22 0.158785252553294
23 0.158271040243614
24 0.158373095582512
25 0.157835649328475
26 0.158340902669908
27 0.158228028238905
28 0.157872019726457
29 0.157953307254791
30 0.158073900689646
31 0.157903330597639
32 0.157768025414794
33 0.157594787595989
34 0.157633947307545
35 0.157642844495659
36 0.157551581325603
37 0.157604162654091
38 0.157542550597934
39 0.157663261143705
40 0.157734187907563
41 0.157645217388955
42 0.157412351570094
43 0.15760508639355
44 0.157629546333484
45 0.157630216943458
46 0.157517112957131
47 0.157597298689631
48 0.15765239033695
49 0.157558954219615
};
\addlegendentry{SSVI (marg.\ lik.)}
\addplot [semithick, red, dashed, forget plot]
table {%
0 0.157229059997287
1 0.158109940411979
2 0.158682414209591
3 0.158322911715549
4 0.158481461718162
5 0.158579070131396
6 0.158505406489465
7 0.157877793264724
8 0.158184744055029
9 0.158311364922765
10 0.157664124775472
11 0.157674232523578
12 0.157428371911785
13 0.157352144203519
14 0.157422228632959
15 0.157510806086004
16 0.157494843666368
17 0.157452602805464
18 0.157459081724498
19 0.15739342741587
20 0.157484269210135
21 0.157394700403719
22 0.157472611321693
23 0.157447116210376
24 0.157492332729681
25 0.157561826687323
26 0.157444550954789
27 0.157405115891984
28 0.157447265144814
29 0.157483528789279
30 0.157524749132419
31 0.157501800537735
32 0.157481186115003
33 0.157510841508215
34 0.157495246173755
35 0.157560862035657
36 0.157542088891988
37 0.157493066769321
38 0.157490531797359
39 0.157526388774786
40 0.157501848916009
41 0.157562507816787
42 0.157507009290162
43 0.157520979962121
44 0.157535939907961
45 0.157496724993198
46 0.157494767438472
47 0.15752984652649
48 0.157538872183548
49 0.15751386251866
};
\addplot [semithick, green!50!black]
table {%
0 0.294961260325784
1 0.182459716115629
2 0.155181499367866
3 0.153276209616279
4 0.156371016386041
5 0.158792923782987
6 0.15992005994062
7 0.160282436242402
8 0.15973821280742
9 0.159244125883517
10 0.159898426103249
11 0.160467708301123
12 0.159990559859941
13 0.159764242751911
14 0.158797503950389
15 0.159056146936802
16 0.159461864770157
17 0.159647660968094
18 0.15922686202361
19 0.159146146535668
20 0.158363316663467
21 0.158689745870812
22 0.158660455975337
23 0.15873547945419
24 0.158206750922481
25 0.158160482653033
26 0.158453073864361
27 0.157975020507363
28 0.157758231265535
29 0.158117001309853
30 0.157736025448876
31 0.158022964836312
32 0.157783014254964
33 0.157742985913377
34 0.157695085500425
35 0.157670830485467
36 0.157698280109862
37 0.157480834041233
38 0.157710298742777
39 0.15770754744148
40 0.157622862256094
41 0.157543569859161
42 0.157689090384515
43 0.157558665790756
44 0.157600879376801
45 0.157608938722606
46 0.157512885866798
47 0.157538498206914
48 0.157618157109679
49 0.157573178395081
};
\addlegendentry{EP}
\addplot [semithick, black, dashed]
table {%
1 0.41250810337944
};
\addlegendentry{Filter init.}
\draw (axis cs:29,0.244806497727642) node[
  scale=0.5,
  anchor=base west,
  text=black,
  rotate=0.0
]{\tikz\node[coordinate] (n2) {};};
\draw (axis cs:-1,0.165) node[
  scale=0.5,
  anchor=base west,
  text=black,
  rotate=0.0
]{\tikz\node[coordinate] (r3) {};};
\draw (axis cs:50,0.152) node[
  scale=0.5,
  anchor=base west,
  text=black,
  rotate=0.0
]{\tikz\node[coordinate] (r4) {};};
\end{axis}

\end{tikzpicture}\\[0em]
    \caption{Test negative log predictive density (NLPD)}
    \label{fig:testing}

    \begin{tikzpicture}[remember picture,overlay]
      \tiny
      \pgfplotsset{yticklabel style={rotate=-90}}
      \setlength{\figurewidth}{.6\textwidth}
      \setlength{\figureheight}{.39\figurewidth}
      \fill[draw=black!50,fill=none] (r3) rectangle (r4);
      \node at (n2) {
\begin{tikzpicture}

\begin{axis}[
height=\figureheight,
tick align=outside,
tick pos=left,
width=\figurewidth,
x grid style={white!69.0196078431373!black},
xmin=-1, xmax=50,
xtick style={color=black},
y grid style={white!69.0196078431373!black},
ymin=0.152, ymax=0.165,
ytick style={color=black}
]
\addplot [semithick, blue]
table {%
0 0.41250810337944
1 0.218024495284717
2 0.1619033682228
3 0.15245697315727
4 0.154281702135226
5 0.156341195644042
6 0.15738839443198
7 0.157643572121716
8 0.157437142487707
9 0.157448874019766
10 0.15757463808453
11 0.157531814750928
12 0.157507029295473
13 0.157752224758334
14 0.157416203021393
15 0.157328205608468
16 0.157482831869204
17 0.157761949558959
18 0.157295073232787
19 0.157991481387726
20 0.157670139793429
21 0.157575397413035
22 0.158155560504331
23 0.157302788450984
24 0.157826355247713
25 0.157321520157085
26 0.157883027529256
27 0.15752230166456
28 0.157588792400429
29 0.157344748183533
30 0.157720086663781
31 0.158050142661836
32 0.158120381068668
33 0.158010566017018
34 0.157582128123029
35 0.157573668826117
36 0.157655396088931
37 0.157480827273648
38 0.15803137998468
39 0.158091333949074
40 0.158137859385754
41 0.157824940936067
42 0.157703577012098
43 0.15758267475298
44 0.157789794076435
45 0.158046247410907
46 0.157556095578216
47 0.157797009991546
48 0.157817607233681
49 0.157737639643816
};
\addplot [semithick, blue, dashed]
table {%
0 0.157229059997197
1 0.15803865241808
2 0.159173727487782
3 0.158510630000205
4 0.158362565955827
5 0.158072805437503
6 0.158648140002203
7 0.158092339890234
8 0.157491630668904
9 0.157861849438426
10 0.157910224517489
11 0.1575656668202
12 0.157847163209446
13 0.157640781662066
14 0.157376133292262
15 0.157406705801527
16 0.157391055037299
17 0.1573965683197
18 0.157347548372605
19 0.157232308209555
20 0.157462204893046
21 0.157355148515592
22 0.157296110882323
23 0.157396036239828
24 0.157365436441189
25 0.157292697210476
26 0.157305140168827
27 0.157313431682967
28 0.157345123018259
29 0.157325201497627
30 0.157364202965764
31 0.157397023455578
32 0.157372240751134
33 0.157333461159887
34 0.15735455700473
35 0.157359912635817
36 0.157418187718972
37 0.157379257461651
38 0.157382351677353
39 0.157408855848361
40 0.157365951313816
41 0.157407604212913
42 0.157355155857192
43 0.157399017952772
44 0.157374837735437
45 0.157386403770081
46 0.157386035817008
47 0.157384228925716
48 0.157377506897062
49 0.15737876017368
};
\addplot [semithick, red]
table {%
0 0.412508103379498
1 0.218165079029874
2 0.162520995452032
3 0.153105488735592
4 0.154854441909495
5 0.157364872981406
6 0.159569722440883
7 0.160348722059811
8 0.159901840583811
9 0.159481506671422
10 0.160670282075018
11 0.159577746437354
12 0.159539250959865
13 0.158914380344364
14 0.159304445737739
15 0.159303730678795
16 0.15946046598674
17 0.158988457568873
18 0.159111187006929
19 0.158495003232138
20 0.159177889212118
21 0.159051390328421
22 0.158785252553294
23 0.158271040243614
24 0.158373095582512
25 0.157835649328475
26 0.158340902669908
27 0.158228028238905
28 0.157872019726457
29 0.157953307254791
30 0.158073900689646
31 0.157903330597639
32 0.157768025414794
33 0.157594787595989
34 0.157633947307545
35 0.157642844495659
36 0.157551581325603
37 0.157604162654091
38 0.157542550597934
39 0.157663261143705
40 0.157734187907563
41 0.157645217388955
42 0.157412351570094
43 0.15760508639355
44 0.157629546333484
45 0.157630216943458
46 0.157517112957131
47 0.157597298689631
48 0.15765239033695
49 0.157558954219615
};
\addplot [semithick, red, dashed]
table {%
0 0.157229059997287
1 0.158109940411979
2 0.158682414209591
3 0.158322911715549
4 0.158481461718162
5 0.158579070131396
6 0.158505406489465
7 0.157877793264724
8 0.158184744055029
9 0.158311364922765
10 0.157664124775472
11 0.157674232523578
12 0.157428371911785
13 0.157352144203519
14 0.157422228632959
15 0.157510806086004
16 0.157494843666368
17 0.157452602805464
18 0.157459081724498
19 0.15739342741587
20 0.157484269210135
21 0.157394700403719
22 0.157472611321693
23 0.157447116210376
24 0.157492332729681
25 0.157561826687323
26 0.157444550954789
27 0.157405115891984
28 0.157447265144814
29 0.157483528789279
30 0.157524749132419
31 0.157501800537735
32 0.157481186115003
33 0.157510841508215
34 0.157495246173755
35 0.157560862035657
36 0.157542088891988
37 0.157493066769321
38 0.157490531797359
39 0.157526388774786
40 0.157501848916009
41 0.157562507816787
42 0.157507009290162
43 0.157520979962121
44 0.157535939907961
45 0.157496724993198
46 0.157494767438472
47 0.15752984652649
48 0.157538872183548
49 0.15751386251866
};
\addplot [semithick, green!50!black]
table {%
0 0.294961260325784
1 0.182459716115629
2 0.155181499367866
3 0.153276209616279
4 0.156371016386041
5 0.158792923782987
6 0.15992005994062
7 0.160282436242402
8 0.15973821280742
9 0.159244125883517
10 0.159898426103249
11 0.160467708301123
12 0.159990559859941
13 0.159764242751911
14 0.158797503950389
15 0.159056146936802
16 0.159461864770157
17 0.159647660968094
18 0.15922686202361
19 0.159146146535668
20 0.158363316663467
21 0.158689745870812
22 0.158660455975337
23 0.15873547945419
24 0.158206750922481
25 0.158160482653033
26 0.158453073864361
27 0.157975020507363
28 0.157758231265535
29 0.158117001309853
30 0.157736025448876
31 0.158022964836312
32 0.157783014254964
33 0.157742985913377
34 0.157695085500425
35 0.157670830485467
36 0.157698280109862
37 0.157480834041233
38 0.157710298742777
39 0.15770754744148
40 0.157622862256094
41 0.157543569859161
42 0.157689090384515
43 0.157558665790756
44 0.157600879376801
45 0.157608938722606
46 0.157512885866798
47 0.157538498206914
48 0.157618157109679
49 0.157573178395081
};
\addplot [semithick, green!50!black, dashed]
table {%
0 0.159313074605827
1 0.159417391260611
2 0.159370933169011
3 0.159091259199324
4 0.158660099637332
5 0.158165743261337
6 0.15827442421702
7 0.158394810767303
8 0.158196899321526
9 0.158198159072234
10 0.157646283858385
11 0.157736542031375
12 0.157510292597481
13 0.157733431524837
14 0.157622637682018
15 0.15753465881844
16 0.157451371705795
17 0.157537434836121
18 0.157564619968464
19 0.157456121027406
20 0.157513052899979
21 0.157431745586963
22 0.157470707244022
23 0.15751111738178
24 0.157529764607361
25 0.157482192666854
26 0.157509098138909
27 0.157521148002931
28 0.157534977909839
29 0.157534847059859
30 0.157500733854391
31 0.157529233057223
32 0.15752019108136
33 0.157518396027358
34 0.157565152668858
35 0.157532150028709
36 0.157553231335575
37 0.157553950924527
38 0.157539398513351
39 0.157520944757576
40 0.157553132031856
41 0.157536995807881
42 0.157561495709671
43 0.157542383468132
44 0.157540276115451
45 0.157563098378081
46 0.157532259416794
47 0.157539569314707
48 0.157569546435478
49 0.157568102414208
};
\draw (axis cs:-1,0.165) node[
  scale=0.5,
  anchor=base west,
  text=black,
  rotate=0.0
]{\tikz\node[coordinate] (z3) {};};
\draw (axis cs:50,0.152) node[
  scale=0.5,
  anchor=base west,
  text=black,
  rotate=0.0
]{\tikz\node[coordinate] (z4) {};};
\end{axis}

\end{tikzpicture}};
      \draw[draw=black!50] (r3)--(z3);
      \draw[draw=black!50] (r4)--(z4); 
    \end{tikzpicture}    
  \end{subfigure}
  \vspace*{-.5em}  
  \caption{Training objective and test performance for various algorithmic choices in the {airline accidents} modelling task, using 10-fold cross validation (mean values shown). The natural gradient parameter updates ensure that SSVI/CVI converges almost as quickly as EP, and convergence can be further sped up by using the filtering (forward) pass for initialization (dashed lines).}

  \label{fig:training-testing}
\end{figure*}

\subsection{Large-scale log-Gaussian Cox Process Modelling}
We now examine the efficacy of the presented SSVI method as a practical machine learning algorithm on a large time series dataset consisting of 1210 dates of commercial airline accidents between 1919--2017 \cite{nickisch2018}. In applying a log-Gaussian Cox process it is necessary to use a bin-width of one day in order to capture the fast varying behaviour (weekly, monthly, and yearly trends), which results in $n = 35{,}959$ observations.

The GP prior contains two components representing long and medium term trends and two representing quasi-periodic behaviour: $\kappa(t,t') = \kappa(t,t')^\text{long}_{\text{Mat.}{\nicefrac{5}{2}}} + \kappa(t,t')^\text{med.}_{\text{Mat.}{\nicefrac{5}{2}}}  + \kappa(t,t')_{\text{Cos}}^{3\,\text{months}} \kappa(t,t')_{\text{Mat.}{\nicefrac{5}{2}}} + \kappa(t,t')_{\text{Cos}}^{1\,\text{week}} \kappa(t,t')_{\text{Mat.}{\nicefrac{5}{2}}} $. We approximate the process with a Poisson likelihood, as above.

\cref{fig:training-testing} shows a comparison of the training objective and test performance for various algorithmic choices. The natural gradient parameter updates ensure that SSVI converges almost as quickly as EP (which is usually posed as a fast-converging alternative to VI). \cref{fig:testing} suggests that using the ELBO as a training objective can result in slower convergence than the marginal likelihood, in terms of test performance.

As discussed in \cref{sec:param_update}, the interpretation of SSVI as a general nonlinear Kalman filter enables us to treat the first filtering pass as an opportunity to \emph{initialise} the variational parameters. Doing so provides a much improved starting point for the optimisation, and results in far superior convergence, as shown by the dashed lines in \cref{fig:training-testing}.

\section{Conclusion}
\label{sec:conclusion}
We have shown how to efficiently employ variational inference in temporal GP models with non-conjugate likelihood models. The method SSVI is a linear-time algorithm that builds on CVI, and is also applicable to more general discrete and continuous-discrete state-space models. We derive the closed-form expressions for efficient evaluation of the variational update step and evidence lower bound (ELBO) by Kalman filtering and smoothing. Furthermore, we proposed an initialization technique for the variational parameters that leverage the forward filter, which showed clear practical benefits. We also demonstrated how to efficiently learn the model hyperparameters using JAX, which allows for automatic differentiation through the state-space model---something that has previously been difficult in major ML frameworks. 

In our experimental validation, we empirically recovered a posterior that matches standard CVI, and demonstrated the benefits of linear-time inference on a large benchmarking problem with around 40 thousand data points. We conclude that JAX shows promise in making auto-differentiation part of the ML toolchain even in sequential models. Codes for this paper are available at \url{http://github.com/AaltoML/kalman-jax}.

\smallskip\noindent
{\bf Author contributions}~~PEC and AS had the original idea after discussions with MEK. WJW and PEC implemented the method and ran the experiments. PEC wrote the first draft of the paper, after which all authors contributed to writing.

\smallskip\noindent
{\bf Acknowledgements}~~This research was supported by grants from the Academy of Finland (grant numbers 308640 and 324345). We acknowledge the computational resources provided by the Aalto Science-IT project.

\section{References}

\begingroup
\small
\setlength{\bibsep}{2pt plus2pt minus2pt}%
\renewcommand{\section}[2]{}%

\bibliographystyle{IEEEbib-abbrev}
\bibliography{bibliography}

\endgroup

\end{document}